\pdfoutput=1
\def\year{2021}\relax
\documentclass[letterpaper]{article} %
\usepackage{aaai21}  %
\usepackage{times}  %
\usepackage{helvet} %
\usepackage{courier}  %
\usepackage[hyphens]{url}  %
\usepackage{graphicx} %
\usepackage{natbib}  %
\usepackage{caption} %
\usepackage{subcaption}
\usepackage{multirow}
\usepackage{booktabs}
\usepackage[switch]{lineno}
\title{A Systematic Evaluation of Object Detection Networks for Scientific Plots}
\author {

        Pritha Ganguly\thanks{The first two authors have contributed equally},
        Nitesh Methani\textsuperscript{*},
        Mitesh M. Khapra,
        Pratyush Kumar \\
}
\affiliations {
    Department of Computer Science and Engineering, \\
    Robert Bosch Centre for Data Science and AI (RBC-DSAI), \\ 
    Indian Institute of Technology Madras, Chennai, India.\\
    \{prithag, nmethani, miteshk, pratyush\}@cse.iitm.ac.in
}
\begin{document}

\maketitle

\begin{abstract}
    \textit{Are existing object detection methods adequate for detecting text and visual elements in scientific plots which are arguably different than the objects found in natural images?} To answer this question, we train and compare the accuracy of Fast/Faster R-CNN, SSD, YOLO and RetinaNet on the PlotQA dataset with over $220,000$ scientific plots. At the standard IOU setting of $0.5$, most networks perform well with mAP scores greater than $80\%$ in detecting the relatively simple objects in plots. However, the performance drops drastically when evaluated at a stricter IOU of 0.9 with the best model giving a mAP of 35.70\%. Note that such a stricter evaluation is essential when dealing with scientific plots where even minor localisation errors can lead to large errors in downstream numerical inferences. Given this poor performance, we propose minor modifications to existing models by combining ideas from different object detection networks. While this significantly improves the performance, there are still two main issues: (i) performance on text objects which are essential for reasoning is very poor, and (ii) inference time is unacceptably large considering the simplicity of plots. To solve this open problem, we make a series of contributions: (a) an efficient region proposal method based on Laplacian edge detectors, (b) a feature representation of region proposals that includes neighbouring information, (c) a linking component to join multiple region proposals for detecting longer textual objects, and (d) a custom loss function that combines a smooth $\ell_1$-loss with an IOU-based loss.
Combining these ideas, our final model is very accurate at extreme IOU values achieving a mAP of 93.44\%@0.9 IOU.
Simultaneously, our model is very efficient with an inference time 16x lesser than the current models, including one-stage detectors.
Our model also achieves a high accuracy on an extrinsic plot-to-table conversion task with an F1 score of 0.77.
With these contributions, we make a definitive progress in object detection for plots and enable further exploration on automated reasoning of plots.
\end{abstract}

\section{Introduction}

\begin{figure}[t]
\centering
\includegraphics[scale=0.18]{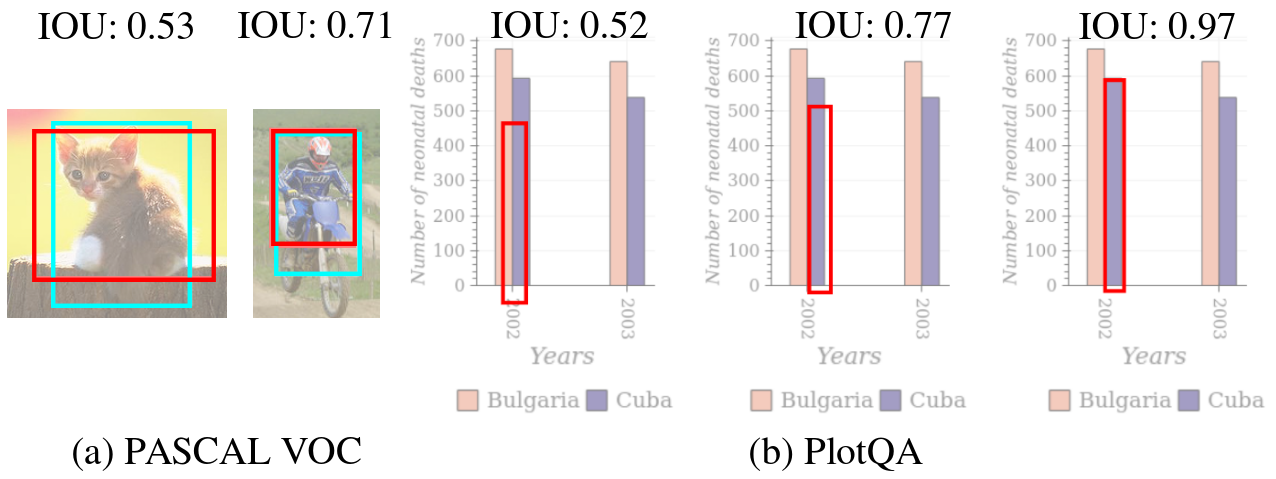}
\caption{Demonstrating sensitivity to IOU on images from (a) PASCAL VOC and (b) PlotQA dataset. Localising on the natural images in (a) is accurate even at low IOU thresholds of 0.5 and 0.75. For the plots in (b), the comparison of plotted values for the two years is incorrect at lower IOU values of 0.5 (left) and 0.75 (centre), and is correct for 0.9 (right).}
\label{fig:voc_vs_plotqa}
\end{figure}

Object detection is one of the fundamental problems in computer vision with the aim of answering \textit{what objects are where} in a given input image. 
Most of the object detection research in the past few years has been on natural images with real-life objects. 
For instance, in the PASCAL VOC dataset \cite{Everingham2010}, the four major classes of objects are people, animals, vehicles, and indoor objects such as furniture.
In this work, we study object detection for a very different class of images, namely computer-generated scientific plots. 
Fig.~\ref{fig:voc_vs_plotqa}b shows an example of a scientific plot: It is a bar plot depicting the number of neonatal deaths in Bulgaria and Cuba over two years. 
Object detection on this plot would be required to identify different visual and textual elements of the plot such as bars, legend previews and labels for ticks, axes \& legend entries.
Such object detection can then enable a question-answering task.
For instance, for Fig.~\ref{fig:voc_vs_plotqa}b we could ask ``In which year does Cuba have lower neonatal deaths?''.
Clearly, this has use-cases in data analytics, and has been studied in recent research \cite{DBLP:conf/cvpr/KaflePCK18,DBLP:conf/iclr/KahouMAKTB18,Methani_2020_WACV}.

It should be clear that scientific plots differ significantly from natural images. 
Firstly, plots have a combination of textual and visual elements interspersed throughout.
The text can either be very short (such as numerical tick labels) or span multiple lines (such as in plot-titles).
Secondly, objects in a plot have a large range of sizes and aspect ratios. 
Depending on the value represented, a bar in a bar-plot can be short or long, while in a line-plot a thin line could depict the data. 
Thirdly, plots impose structural relationships between some of the objects. 
For instance, the legend preview and the close-by title text denote a correspondence map. 
This also applies to a tick label and its corresponding bar in a bar-plot. 

Given these differences, it needs to be seen if existing object detection methods are adequate for scientific plots. 
In particular, are they capable of (a) detecting short and long pieces of text, (b) detecting objects with large data-dependent range of sizes and aspect ratios, and (c) localising objects accurately enough to extract structural relationships between objects?
To answer this question, we first evaluate state-of-the-art object detection networks on the PlotQA dataset \cite{Methani_2020_WACV} which has over $220,000$ scientific plots sourced from real-world data thereby having a realistic vocabulary of axes variables and complex data patterns which mimic the plots found in scientific documents and reports. 
We observe that, across these models, the average of mAP@$0.5$ is only around $87\%$, indicating success in detecting the relatively simple objects in the plot.

While the above results appear positive, a closer manual inspection revealed that these models make critical errors which lead to large errors in downstream numerical inference on the plots. 
This disparity is because we use an IOU of 0.5 while computing the mAP scores. 
While IOU values in the range of 0.5 and 0.7 are acceptable for natural images where large relative areas are covered by foreground objects, such values are unacceptably low for scientific plots.
This is demonstrated in Fig.~\ref{fig:voc_vs_plotqa}a with two example images from the PASCAL VOC dataset where the predicted box (red) is very different from the ground-truth box (cyan), but still acceptable as the IOU is within range. 
Contrast this with the case for the plot in Fig.~\ref{fig:voc_vs_plotqa}b. 
For an IOU setting of 0.5 (left) and 0.75 (middle), the estimated values of the data-points would incorrectly identify that neonatal deaths in Cuba are lower in 2002 than the actual value in 2003. 
Only at the high IOU value of 0.9, this is correctly resolved.
Thus, downstream numerical reasoning on plots requires much stricter IOU settings in comparison to object detection on natural images. 
For the PlotQA dataset if we use a stricter IOU of 0.9, then the mAP scores for all models drop drastically with the best model giving a mAP of only $35.70\%$. 
In particular, one-stage detectors such as SSD \cite{DBLP:conf/eccv/LiuAESRFB16} and YOLO-v3 \cite{DBLP:journals/corr/abs-1804-02767} have a single-digit mAP@0.9.

The poor performance of current models at high IOU settings motivate improvements in the models. We first propose minor modifications to existing models. In particular, we propose a hybrid network which combines Faster R-CNN \cite{DBLP:conf/nips/RenHGS15} with the Feature Pyramid Network \cite{DBLP:conf/cvpr/LinDGHHB17} from RetinaNet \cite{DBLP:conf/iccv/LinGGHD17} and the ROIAlign idea from Mask R-CNN \cite{DBLP:conf/iccv/HeGDG17}. This significantly improves the performance giving an overall mAP of $77.22$\%@0.9 IOU. However, careful analysis reveals two major limitations: (i) accuracy on text objects is very low which can lead to errors in downstream analytics tasks, and (ii) the inference time is very high (374 ms) which is unacceptable given the lower visual complexity of plots.

To further improve the speed and performance, we propose an architecture, named PlotNet, which contains (i) a fast and conservative region proposal method based on Laplacian edge detectors, (ii) an enhanced feature representation of region proposals that includes local neighbouring information, (iii) a linking component that combines multiple region proposals for better detection of longer textual objects, and (iv) a custom regression loss function that combines smooth $\ell_1$-loss with an IOU-based loss designed for improving localisation at higher IOUs. This significantly improves accuracy with a mAP of 93.44\%@0.9 IOU. Further, it is 16x faster than its closest competitor and has 3x lower FLOPs.
We also evaluate PlotNet on an extrinsic plot-to-table conversion task where we extract the plot's underlying data into a table.
Specifically, we replace Faster R-CNN with PlotNet in the Visual Element Detection stage of the pipeline proposed in PlotQA \cite{Methani_2020_WACV}. This results in a relative improvement of 35.08\% in the table F1 score.

In summary, the contributions of this paper are as follows:
\begin{enumerate}
    \item We motivate the need for object detection at extreme IOU values for the specific dataset of scientific plots which require accurate localization.
    \item We evaluate the robustness of nine different object detection networks (SSD, YOLO-v3, RetinaNet, variants of Fast and Faster R-CNN) to increase in IOU and identify that Feature Pyramid Network (FPN) and ROIAlign (RA) are good design choices for higher accuracy.
    \item We propose PlotNet that improves on mAP by over $16$ points while reducing execution time by over $16$ times from its closest competitor. Thus, PlotNet is faster than one-stage detectors and simultaneously more accurate than the best two-stage detectors.
\end{enumerate}

The rest of the paper is organised as follows: We first discuss the different datasets and detail the experimental setup for evaluating existing object detection models on scientific plots. We also report the results and critique their performance on an example plot image from the PlotQA dataset. We then detail the architecture of PlotNet and compare it with other networks in terms of accuracy and speed. Note that, we follow an unconventional organisation of describing partial experimental results first as this lays the groundwork for motivating the design and evaluation of PlotNet.

\section{Evaluation of Existing Models}

\begin{table*}[ht!]
\footnotesize
\begin{tabular}{@{}clcccccccccccc@{}}
\toprule
 &  & \multicolumn{10}{c}{\textbf{IOU @0.9}} & \multicolumn{1}{l}{\textbf{@0.75}} & \multicolumn{1}{l}{\textbf{@0.5}} \\ \midrule
\textbf{S.L} & \multicolumn{1}{c}{\textbf{Existing Models}} & \textbf{bar} & \textbf{\begin{tabular}[c]{@{}c@{}}dot-\\ line\end{tabular}} & \textbf{\begin{tabular}[c]{@{}c@{}}legend\\ label\end{tabular}} & \textbf{\begin{tabular}[c]{@{}c@{}}legend\\ preview\end{tabular}} & \textbf{\begin{tabular}[c]{@{}c@{}}plot\\ title\end{tabular}} & \textbf{\begin{tabular}[c]{@{}c@{}}x-axis\\ label\end{tabular}} & \textbf{\begin{tabular}[c]{@{}c@{}}x-axis\\ ticks\end{tabular}} & \textbf{\begin{tabular}[c]{@{}c@{}}y-axis\\ label\end{tabular}} & \textbf{\begin{tabular}[c]{@{}c@{}}y-axis\\ ticks\end{tabular}} & \textbf{mAP} & \textbf{mAP} & \textbf{mAP} \\ \midrule 
(a) & SSD & 1.39 & 0.60 & 2.18 & 0.39 & 0.04 & 3.39 & 0.44 & 5.14 & 0.20 & 1.53 & 39.78 & 82.33 \\
(b) & YOLO-v3 & 15.51 & 8.72 & 7.15 & 11.70 & 0.02 & 4.39 & 8.08 & 9.59 & 1.70 & 7.43 & 73.31 & 96.27 \\
(c) & RetinaNet & 16.51 & 18.5 & 77.26 & 29.74 & \textbf{16.58} & 67.62 & 28.40 & 3.14 & 17.31 & 30.56 & 81.13 & 90.13 \\
(d) & FRCNN & 53.38 & 1.68 & 12.59 & 14.06 & 0.03 & 42.13 & 25.49 & 11.68 & 31.98 & 21.45 & 63.68 & 72.83 \\
(e) & FrRCNN & 6.92 & 1.68 & 1.39 & 1.45 & 0.00 & 4.35 & 6.10 & 3.57 & 5.18 & 4.08 & 50.51 & 88.49 \\
(f) & Mask R-CNN & 47.54 & 5.36 & 50.83 & 32.43 & 0.33 & 40.20 & 33.72 & 80.53 & 30.31 & 35.70 & 82.45 & 93.72 \\ \midrule 
(g) & FRCNN (FPN+RA) & \textbf{87.59} & \textbf{31.62} & 79.05 & 66.39 & 0.22 & 69.78 & 88.29 & 46.63 & 84.60 & 61.57 & 69.82 & 72.18 \\
(h) & FrRCNN (RA) & 63.89 & 14.79 & 70.95 & 60.61 & 0.18 & 83.89 & 60.76 & 93.47 & 50.87 & 55.49 & 89.14 & 96.80 \\
(i) & FrRCNN (FPN+RA) & 85.54 & 27.86 & \textbf{93.68} & \textbf{96.30} & 0.22 & \textbf{99.09} & \textbf{96.04} & \textbf{99.46} & \textbf{96.80} & \textbf{77.22} & \textbf{94.58} & \textbf{97.76} \\ \bottomrule
\end{tabular}
\caption{Comparison of existing and hybrid models on the PlotQA dataset with mAP scores (in \%) at IOUs of 0.9, 0.75, and 0.5. For IOU@0.9, the class-wise average precision (in \%) is shown for all the classes.}
\label{tab:existing_models_map}
\end{table*}

\noindent\textbf{Dataset: }Automated visual analysis and subsequent question-answering on scientific plots was first proposed in FigureQA \cite{DBLP:conf/iclr/KahouMAKTB18}.
There are three publicly available datasets, namely FigureQA \cite{DBLP:conf/iclr/KahouMAKTB18}, DVQA \cite{DBLP:conf/cvpr/KaflePCK18}, and PlotQA \cite{Methani_2020_WACV}.
These datasets contain scientific plots with bounding boxes and labels for different plot elements including bars, lines, tick labels, legend entries, and plot labels. 
We run our experiments on the PlotQA dataset \cite{Methani_2020_WACV}, as it is based on real-world data while both FigureQA and DVQA are based on synthetic data. 
For instance, in synthetic datasets, the label names are selected from a limited vocabulary such as colour names in FigureQA and top-1000 nouns in the Brown corpus in DVQA. 
On the other hand, PlotQA has datasets collected from public data repositories. 
This impacts object detection as the text labels show large variability in PlotQA dataset.
Secondly, synthetic datasets use limited range of plotted values such as integers in a fixed range, while PlotQA plots real data. 
This impacts object detection as the size of bars in a bar-plot and the slopes in a line-plot show large variability in the PlotQA dataset. 

The PlotQA dataset \cite{Methani_2020_WACV} contains over $220,000$ scientific plots across three categories of bar (both horizontal and vertical), line, and scatter plots. 
The dataset includes ground-truth bounding boxes for bars, dot-lines, legend labels, legend previews, plot-title, axes ticks and their labels.
The underlying data is from data sources such as World Bank Open Data containing natural variable names such as mortality rate, crop yield, country names, and so on.

\noindent\textbf{Training Setup:}
We used the existing implementations for the R-CNN family, YOLO-v3, SSD and RetinaNet. 
ResNet-50 (R-50) pre-trained on ImageNet \cite{DBLP:conf/cvpr/DengDSLL009} dataset is the backbone feature extractor for Fast R-CNN, Faster R-CNN, Mask R-CNN and RetinaNet. 
For SSD and YOLO-v3, InceptionNet \cite{DBLP:conf/cvpr/SzegedyLJSRAEVR15} and DarkNet-53 were the backbone feature extractors, respectively.

These models were trained with %
an initial base learning rate of $0.025$ with momentum stochastic gradient descent algorithm.
The network's classification and regression heads use a batch-size of $512$ ROIs.
RetinaNet and SSD models were trained with a batch-size of $32$ with a learning rate of $0.004$. %
Based on evaluation on the validation dataset, we modified the parameters in the focal loss for RetinaNet to $\alpha=0.75$ and $\gamma=1.0$, against recommended values of $\alpha=0.25$ and $\gamma=1.0$. 
The model was trained with a batch-size of $64$ and a learning rate of $0.001$.%

\noindent{\textbf{Results \& Comparative Analysis}}: For each of the models, mAP score at three different IOU values of 0.9, 0.75, and 0.5 are shown in Table~\ref{tab:existing_models_map}.
Here are the important observations: 
\begin{itemize}
    \item mAP@0.5 is fairly high (average over $87\%$) across models indicating that the relatively simple visual elements of the plots are being identified with high accuracy.
    \item mAP@0.75 falls markedly in comparison to mAP@0.5, with an average drop of about 22 points across models. 
    Specifically for SSD, Faster R-CNN the drop is very high at about 40 points. 
    \item mAP@0.9 drops to remarkably low values; on average mAP@0.9 is less than half of mAP@0.75.
    Specifically, SSD, Faster R-CNN, and YOLO have single-digit mAP@0.9 values. 
    \item For an IOU of 0.9, the AP for individual classes shows large variability across models.
    Relatively, plot-title and dot-line classes have the lowest AP values across models.
\end{itemize}

\begin{figure*}[t]
\centering
\includegraphics[width=0.95\linewidth]{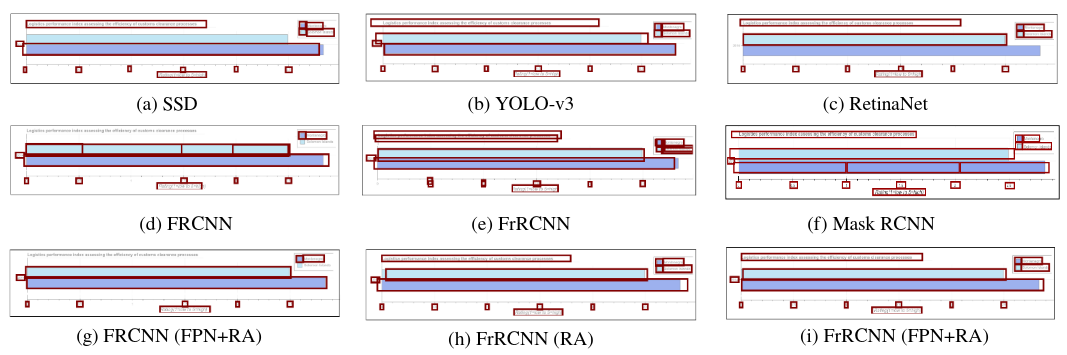}
\caption{Detected bounding boxes on an example plot from PlotQA for different models corresponding to Table~\ref{tab:existing_models_map} at IOU 0.9.}
\label{fig:all_detections}
\end{figure*}
To better illustrate the performance of each model, we exemplify the bounding box outputs of the different models on specific parts of an example plot shown in Fig.~\ref{fig:all_detections}.
We make the following observations, model-wise. 
\begin{itemize}
    \item \textbf{SSD} glaringly misses detecting one of the bars, and also has low localisation accuracy as evidenced in the misaligned bounding for the second bar in Fig.~\ref{fig:all_detections}a.
    However, it correctly detects small tick labels, perhaps due to proposal generation performed at multiple resolutions. 
    \item \textbf{YOLO-v3} detects all objects (including both bars), but with lower localisation accuracy. 
    For instance in Fig.~\ref{fig:all_detections}b, the upper bar and plot title have misaligned bounding boxes.
    To see if this problem could have been solved by imposing priors on aspect ratios of bounding boxes, we plotted aspect ratios of all objects across plots and found no distinct clusters, \textit{i.e.}, aspect ratios of bars, texts, \textit{etc.} vary in a continuum. This makes it hard to choose appropriate priors for bounding boxes.
    \item \textbf{RetinaNet} which is based on SSD also misses out on detecting one of the bars and also the y-tick label (Fig.~\ref{fig:all_detections}c).
    The bounding box of the detected bar is more accurate than that in SSD, indicating the benefit of the lateral connections in generating features for the regression head.
    Across the three one-stage detectors, which have much higher speed than the two-stage detectors, RetinaNet is the clear winner (row (c) in Table~\ref{tab:existing_models_map}).
    While not apparent in the illustrated example, RetinaNet's focal loss with custom tuned parameters $(\alpha, \beta)$ instead of hard suppression, may also be contributing to its higher performance. 
    \item \textbf{Fast R-CNN} (FRCNN) breaks up one of the bars into smaller objects (aligning with lines on the background grid of the plot). 
    It also misses several objects including a legend item, the plot title, and a tick label (Fig.~\ref{fig:all_detections}d). 
    This could be attributed to the proposal generation method which uses selective search (SS).
    This under-performance is also visible at low IOUs: mAP@0.5 is lowest for Fast R-CNN (row (d) in Table~\ref{tab:existing_models_map}) potentially due to poorly performing SS which remains unaffected by IOU. 

    \item \textbf{Faster R-CNN} (FrRCNN) improves over the recall of SS by detecting most objects due to more complex region proposal network (RPN). 
    However, RPN creates multiple overlapping proposals, even after non-maximal suppression (NMS) (Fig.~\ref{fig:all_detections}e). 
    This lowers the bulky model's mAP to just $4\%$, which is second-lowest.
    \item \textbf{Mask R-CNN} uses Faster R-CNN as the backbone architecture but uses ROIAlign instead of ROIPool gives mixed results when compared to Faster R-CNN. It is able to detect the longer textual elements (\textit{e.g}., title) but has poorer localisation accuracy on the bars. It also breaks up one of the bars into smaller objects. However, its localisation accuracy on the tick labels is better than Faster R-CNN. 
\end{itemize}
In summary, most state-of-the-art models have low robustness to high IOU values on this different class of images.

\noindent\textbf{A hybrid network combining existing ideas: }
The above discussion clearly establishes the need for better models for object detection over scientific plots. However, before we do so, we wanted to examine if combining ideas from existing models can help in improving the performance. Among the one-stage detectors, RetinaNet (row(c) in Table~\ref{tab:existing_models_map}) gave the best performance. Similarly, among the two-stage detectors Mask R-CNN (row(f) in Table~\ref{tab:existing_models_map}) gave the best performance. However, the qualitative analysis presented in Fig.~\ref{fig:all_detections} suggested that Faster R-CNN has some advantages over Mask R-CNN. Taking all of this into consideration we decided to combine the relative merits of Faster R-CNN, RetinaNet and Mask R-CNN. In particular, we retain the overall architecture of Faster R-CNN but use FPN as the feature extractor (as in RetinaNet) and replace ROIPool with ROIAlign (as in Mask R-CNN). The results obtained by making these modifications are summarised in rows (g), (h) and (i) of Table~\ref{tab:existing_models_map}. For the sake of completeness we also present results obtained by combining Fast R-CNN (FRCNN) with FPN and ROIAlign. We observe that across all three IOU values, the highest mAP values are obtained by Faster R-CNN with FPN and ROIAlign (row(i) in Table~\ref{tab:existing_models_map}). This mAP of 77.22\% is the best number reported in Table~\ref{tab:existing_models_map}. 

We do the same qualitative analysis as before for the three hybrid models and make the following observations:

\begin{itemize}
    \item \textbf{Fast R-CNN with FPN and ROIAlign} (FRCNN (FPN+RA)) improves on Fast R-CNN by not breaking up the bar into smaller objects, due to the FPN which enables improved feature extraction for the regression head. 
   However due to the use of selective search (SS), many objects continue to go undetected (Fig.~\ref{fig:all_detections}g). 
    Notably, for the detected objects, the localisation accuracy is high, perhaps due to the use of ROIAlign's bilinear interpolation when mapping proposals into smaller cells.
    \item \textbf{Faster R-CNN with ROIAlign} (FrRCNN (RA)) improves on Faster R-CNN due to the substitution of ROIPool with ROIAlign. 
    This leads to more accurate localisation and thereby removal of multiple proposals by NMS (Fig.~\ref{fig:all_detections}h).
    Interestingly, this model under-performs FRCNN (FPN+RA) as evident in the mAP values and in the example by the difference in localisation accuracy.  
    This illustrates the importance of FPN in being able to nullify the limitations of SS and improve localisation. 
    \item \textbf{Faster R-CNN with FPN and ROIAlign} (FrRCNN (FPN+RA)) performs the best amongst all existing models. 
    It combines the RPN of Faster R-CNN for better region proposals, with FPN which provide better features, and ROIAlign which provides better mapping of features to scaled cells (Fig.~\ref{fig:all_detections}i).
    There is an additive effect of combining these three important ideas in object detection, as evidenced by the significant difference between this model and the rest.%
\end{itemize}
In summary, the model with the highest performance is the one that combines the best ideas in object detection.

\section{Our Proposed Model}
\label{sec:our_model}

\begin{figure*}[ht!]
\centering
\includegraphics[scale=0.27]{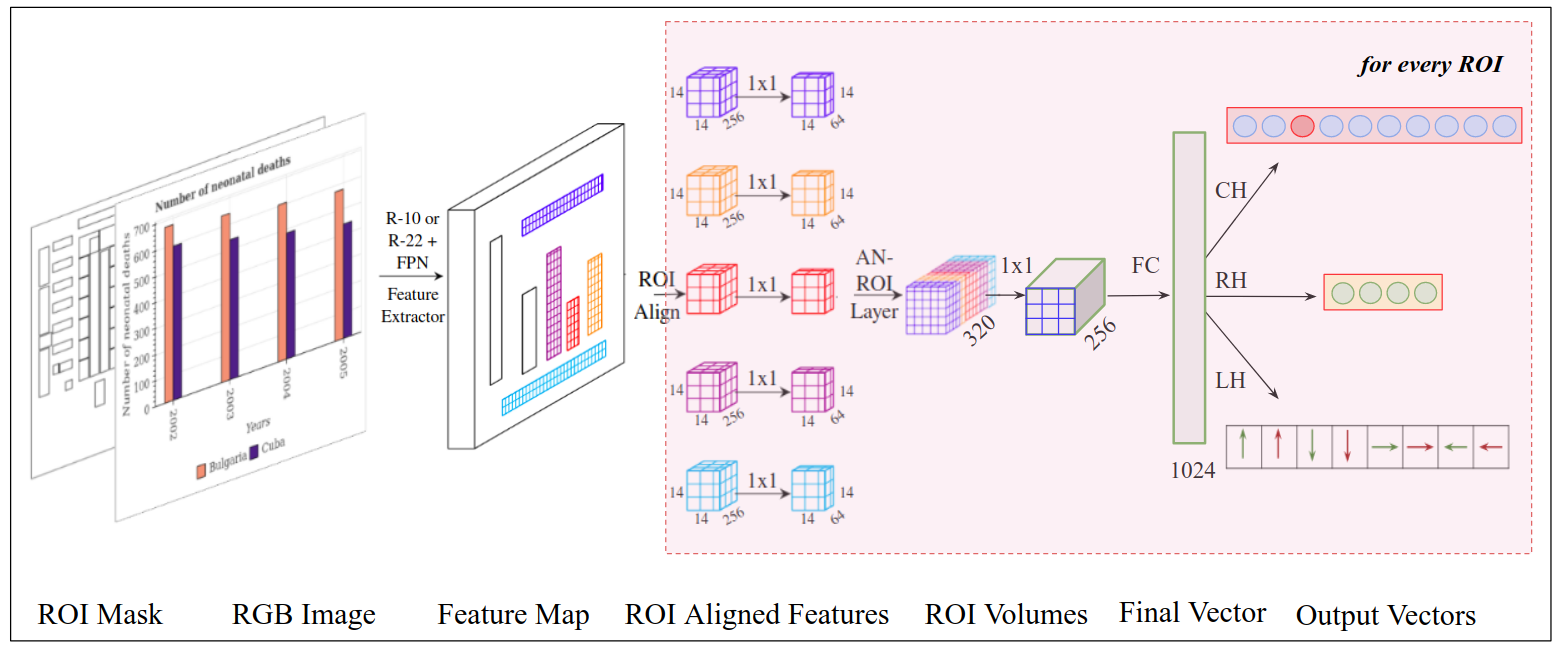}
\caption{Architecture diagram of our proposed model, PlotNet.}
\label{fig:plotnet_arch}
\end{figure*}

Given the low mAP scores of existing models at the requisite IOU of 0.9, we propose a new network (shown in Fig.~\ref{fig:plotnet_arch}) which is designed bottom up based on three key observations. 
First, we observe that networks which use existing region proposal methods such as selective search, RPN, and anchor based methods have low mAP@0.9.
In particular, these methods either generate too many proposals or miss out some objects. 
We contrast this with the apparent low visual complexity of these plots which suggests that detecting region proposals should be easier. 
Based on this insight, we propose a region proposal method which relies on traditional CV-based methods such as edge detection followed by contour detection. 
We however retain ROIAlign and FPN components which improved the performance of models. 
However, we note that FPN adds a significant computational cost and its addition needs to be carefully evaluated. 
While doing so, we make our second observation that longer textual elements such as titles and legend labels get detected as multiple proposals which need to be linked. 
We propose a separate linking component which decides whether a given proposal needs to be merged with any of its neighbours. 
None of the existing models perform such linking. 
Third, we notice a sharp decline of mAP scores on increase in the IOU. 
To address this, we design a custom loss function, which has non-negligible loss values for high IOU ( $> 0.8$).

In summary, in the design of a custom model which we refer to as PlotNet, we (i) use a computationally efficient CV-based region proposal method, (ii) supplement it with a link prediction method to detect contiguous text objects, (iii) use ROIAlign and neighbouring information to better map proposals into smaller cells, (iv) evaluate the necessity of FPN, and (v) evaluate the need for IOU-based loss functions.
We now describe the different components of our model.

\noindent\textbf{Region Proposal:} 
As an alternative to Region Proposal Network (RPN), we propose a combination of CV methods to generate region proposals (Fig.~\ref{fig:proposals_loss_comparison_plotnetv7}(a)). 
Specifically, we (a) draw edges with a Laplacian edge detector on the image of the plot, (b) extract continuous closed contours from edges based on uniform color and/or intensity, (c) convert contours to a bounding-box by finding the minimal up-right bounding rectangle for each of the identified contours. These boxes serve as regions of interest (ROI) which are passed as input to the network. 
These proposals are very small in number ($90$ proposals on an average) in comparison to selective search which gives around $2k$ proposals. 
Further, while selective search takes $\sim 6740$ms per image on average to generate proposals, our method takes only $34$ms. We refer the readers to the supplementary section for more details.

\noindent\textbf{Feature Extraction:}
We use ResNet \cite{DBLP:conf/cvpr/HeZRS16} for extracting features from the input image after resizing the image to $650 \times 650$. %
To exploit structural information present in the image, we add the ROIs proposed earlier as the $4^{th}$ channel to the RGB input image. 
We tried different number of layers in the ResNet model \& found that even with 10 layers we were able to get a good performance. 
We also consider FPN as the feature extractor with ResNet-22 as the backbone architecture as a potential trade-off between performance and cost.
Once the image features are extracted, the externally generated ROIs are projected onto the feature map. 
To extract a fixed sized ROI feature, we pass them through the ROIAlign layer \cite{DBLP:conf/iccv/HeGDG17} which outputs the fixed size feature map of size $14 \times 14 \times 256$. 
We further reduce the depth of each ROI feature map to $14 \times 14 \times 64$ by using $1 \times 1$ convolution layers.

\noindent \textbf{Accumulating Neighbouring ROI Information (AN-ROI layer):} 
To incorporate local neighbouring information into each ROI feature, we create an AN-ROI volume by concatenating the ROI's immediate left, right, top and bottom features along the depth, resulting in a feature volume of size $14 \times 14 \times 320$. 
We then apply convolutional layers on this AN-ROI volume resulting in a feature volume of size $14 \times 14 \times 256$.
We hypothesise that such neighbouring features would increase the accuracy of classifying, regressing, or linking individual ROIs.

\noindent\textbf{Classification (CH), Regression (RH) \& Linking heads (LH):} 
The ROI features extracted above are passed through two fully-connected layers with $2048$ and $1024$ neurons, respectively. 
Each ROI feature vector, is independently passed through the CH which uses the softmax function to output a probability distribution over the $9$ classes of objects in the images and a tenth background class. 
These $9$ classes are bar, dot-line, legend-preview, legend-label, plot-title, x-axis label, y-axis label, x-axis ticks and y-axis ticks. 
The same ROI feature vector is also fed to the RH which predicts the $4$ co-ordinates (top-left and bottom-right) of the bounding box. 
Lastly, the same ROI feature vector is passed to the LH which predicts whether this ROI needs to be merged with none, one, or more of its immediate 4 (top, left, right and bottom) neighbours.

\noindent{\textbf{IOU-based Loss functions}}:
\begin{figure}[t]
\centering
\includegraphics[scale=0.23]{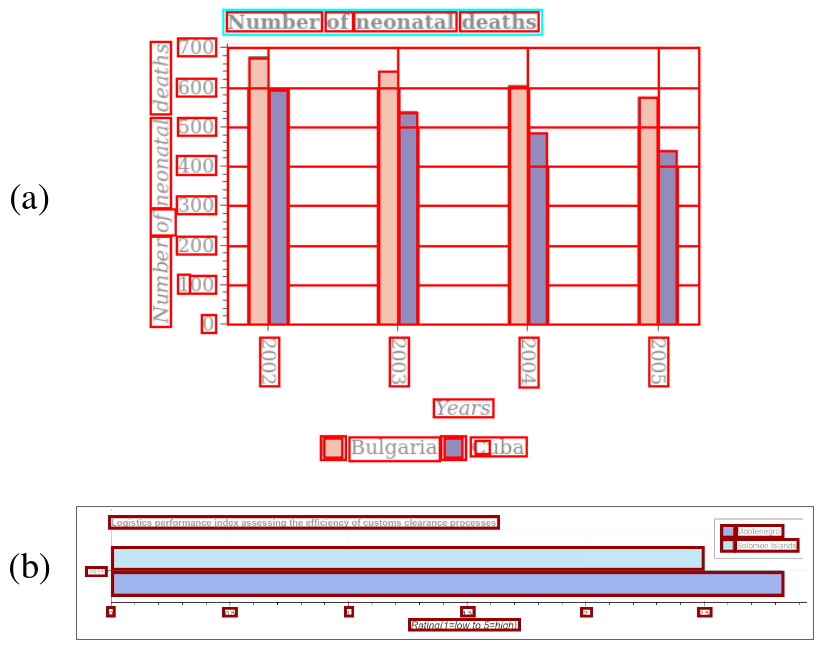}
\caption{(a) The proposals generated by our CV method are shown in red. (b) Detected bounding boxes by PlotNet on an example plot from PlotQA dataset at IOU 0.9.} 
\label{fig:proposals_loss_comparison_plotnetv7}
\end{figure}
Most state-of-the-art object detection models use $\ell_n$ loss (\textit{e.g.}, $\ell_2$-loss, smooth $\ell_1$-loss) for performing bounding box regression. However, several studies \cite{DBLP:conf/mm/YuJWCH16, DBLP:conf/cvpr/BermanTB18} suggest that there are some disadvantages of doing so and instead an IOU-based loss function which better correlates with the final evaluation metric should be used. Indeed, some studies have showed that using $-\log IOU$ \cite{DBLP:conf/mm/YuJWCH16} and $1-IOU$ \cite{DBLP:conf/cvpr/BermanTB18} as loss functions give better results by ensuring that (i) the training objective is aligned with the evaluation metric, and (ii) all the 4 coordinates of the bounding box are considered jointly. 
However, these loss functions fail to learn anything in case of non-overlapping boxes. To overcome this, more generalized loss functions like GIOU \cite{DBLP:conf/cvpr/RezatofighiTGS019}, DIOU, and CIOU \cite{DBLP:conf/aaai/ZhengWLLYR20} have been proposed which add an additional penalty term to deal with the non-overlapping boxes.

In existing loss functions, the penalty is negligible for boxes which have a large IOU overlap with the ground-truth box. To enable the network to learn tighter bounding boxes, motivated by focal loss \cite{DBLP:conf/iccv/LinGGHD17}, we propose a Focal IOU (FIOU) loss that more gradually decreases the penalty as IOU overlap with ground-truth box increases.
Formally, FIOU is defined as:
$$ \mathcal{L}_{FIOU} = -(1+IOU)^\gamma * \log(IOU)$$
FIOU focuses on higher IOU values by dynamically scaling the $\log(IOU)$ loss \cite{DBLP:conf/mm/YuJWCH16}, where the scaling factor increases as the IOU increases. The higher the hyperparameter $\gamma$, the larger is the penalty at medium IOU levels. We experimented with a couple of $\gamma$ values and have found  $\gamma=2$ to work best for the PlotQA dataset. We include a plot comparing these loss functions graphically in the supplementary section.
We then define a custom loss function combining smooth $\ell_1$-loss (SL1) with FIOU: $\mathcal{L}_{Custom} = \mathcal{L}_{SL1} + \mathcal{L}_{FIOU}$. This custom loss function achieves state-of-the-art results as we report in the comparison across loss functions in Table~\ref{tab:plotnet_map}.

\noindent{\textbf{Training \& Implementation Details}}: While training, for every proposed ROI, we need to assign a ground-truth class for the 9 object classes and the background class.
We identify the ROI's center and identify if it lies in any of the ground-truth bounding boxes. 
There would be at most one such box, since objects in plots do not overlap unlike natural images. 
If no such ground-truth bounding box is found, the ROI is considered to be in the background class and is ignored by the regression head.

Similarly, for every proposed ROI, the coordinates of the parent ground-truth box identified above are assigned as the regression targets. 
In particular, for visual ROIs,
the regression target is set to the co-ordinates of the parent box. 
For textual objects,
it is difficult to regress the ROI to match the entire span of the parent box. 
For example, in Fig.~\ref{fig:proposals_loss_comparison_plotnetv7}(a), for the ROI containing the word ``Number'' in the title, the ground-truth box would be the entire title spanning all the words (cyan box).
To avoid this large difference from the proposal, we create the regression targets for ``Number'' by clipping the ground-truth box to have the same boundary as the proposed box along the horizontal direction. 
The task then is to grow the proposed ROI vertically and then later link it to its neighbour thereby creating the entire title box.

Lastly, for creating the ground-truth for linking, we assign a binary value to each ROI for each of the 4 directions. 
These 4 values indicate whether the ROI needs to be linked to its left, right, top, or bottom neighbours. 
In order to find the neighbours, we consider an area of $50\times50$ around a ROI and check if any of the neighbouring ROIs have the same parent box. 
If so, we assign 1 to the link corresponding to the direction (top, right, bottom, left) of that neighbour.

We trained our model for $10$ epochs using Adam optimizer \cite{article} with a learning rate of $0.0001$. We experimented with different loss functions for bounding box regression, and used cross-entropy loss for classification as well as link prediction.

\section{Discussion}
We now discuss the performance of different variants of our model as reported in Table~\ref{tab:plotnet_map}. Note that all the variants in Table~\ref{tab:plotnet_map} use FPN as we always get better results with FPN. The variants mentioned in rows (a) and (b) use smooth $\ell_1$-loss (SL1) as the regression loss and do not have the Linking Head (LH) and the AN-ROI layer, respectively. The variants mentioned in rows (c) to (j) comprise the LH and the AN-ROI layer but only differ with respect to the regression loss.
We make the following observations:

\begin{table*}[t!]
\footnotesize
\begin{tabular}{@{}clcccccccccccc@{}}
\toprule
 &  & \multicolumn{10}{c}{\textbf{IOU @0.9}} & \multicolumn{1}{l}{\textbf{@0.75}} & \multicolumn{1}{l}{\textbf{@0.5}} \\ \midrule
\textbf{S.L }& \multicolumn{1}{c}{\textbf{PlotNet Variants}} & \textbf{bar} & \textbf{\begin{tabular}[c]{@{}c@{}}dot-\\ line\end{tabular}} & \textbf{\begin{tabular}[c]{@{}c@{}}legend\\ label\end{tabular}} & \textbf{\begin{tabular}[c]{@{}c@{}}legend\\ preview\end{tabular}} & \textbf{\begin{tabular}[c]{@{}c@{}}plot\\ title\end{tabular}} & \textbf{\begin{tabular}[c]{@{}c@{}}x-axis\\ label\end{tabular}} & \textbf{\begin{tabular}[c]{@{}c@{}}x-axis\\ ticks\end{tabular}} & \textbf{\begin{tabular}[c]{@{}c@{}}y-axis\\ label\end{tabular}} & \textbf{\begin{tabular}[c]{@{}c@{}}y-axis\\ ticks\end{tabular}} & \textbf{mAP} & \textbf{mAP} & \textbf{mAP} \\ \midrule 
(a) & No LH & 85.30 & 52.85 & 29.64 & 94.30 & 0.00 & 10.36 & 80.77 & 1.47 & 81.59 & 48.48 & 53.71 & 54.03 \\
(b) & No AN-ROI & 91.02 & 31.69 & 97.08 & 81.57 & 99.36 & 96.06 & 85.33 & 82.00 & 90.95 & 83.89 & 97.21 & 98.11 \\ \midrule 
(c) & $\mathcal{L}_{SL1}$ & 92.16 & 61.18 & 98.38 & 93.46 & 99.44 & 97.21 & 94.21 & 95.45 & 94.42 & 91.77 & 97.74 & 98.24 \\
(d) & $\mathcal{L}_{1 - IOU}$ & 91.79 & 41.86 & 93.74 & 94.64 & 98.29 & 83.11 & 85.69 & 89.32 & 49.36 & 80.87 & 96.38 & 98.20 \\
(e) & $\mathcal{L}_{-\log IOU}$ & 91.83 & 45.78 & 91.48 & 94.15 & 98.95 & 74.24 & 87.19 & 89.34 & 50.11 & 80.34 & 96.97 & 98.26 \\
(f) & $\mathcal{L}_{GIOU}$ & 91.71 & 49.30 & 95.99 & 93.55 & 98.42 & 95.03 & 89.77 & 94.08 & 86.06 & 88.21 & 96.37 & 98.16 \\
(g) & $\mathcal{L}_{DIOU}$ & 91.35 & 52.22 & 96.31 & 93.45 & 96.82 & 96.18 & 89.63 & 95.46 & 94.07 & 89.50 & 97.17 & 98.22 \\
(h) & $\mathcal{L}_{CIOU}$ & 91.15 & 55.03 & 97.89 & 92.99 & 99.46 & 96.33 & 91.30 & 90.40 & \textbf{95.48} & 90.00 & 97.27 & 98.28 \\
(i) & $\mathcal{L}_{FIOU}$ & 91.88 & 61.44 & 96.44 & 95.58 & 99.27 & 97.19 & 90.64 & 97.55 & 87.66 & 90.88 & 97.30 & 98.31 \\ \midrule 
(j) & $\mathcal{L}_{Custom}$ & \textbf{92.80} & \textbf{70.11} & \textbf{98.47} & \textbf{96.33} & \textbf{99.52} & \textbf{97.31} & \textbf{94.29} & \textbf{97.66} & 94.48 & \textbf{93.44} & \textbf{97.93} & \textbf{98.32} \\ \bottomrule
\end{tabular}
\caption{Comparison of variants of PlotNet on the PlotQA dataset with mAP scores (in \%) at IOUs of 0.9, 0.75, and 0.5.}
\label{tab:plotnet_map}
\end{table*}

\begin{table}[t!]
\footnotesize
\centering
\begin{tabular}{lccc}
\hline
\multicolumn{1}{c}{\textbf{Models}} & \textbf{Precision} & \textbf{Recall} & \textbf{F1-score} \\ \hline
FRCNN (FPN+RA) & 0.63 & 0.02 & 0.04 \\
FrRCNN (RA) & 0.62 & 0.12 & 0.20 \\
FrRCNN (FPN+RA) & 0.62 & 0.52 & 0.57 \\
PlotNet - Ours & \textbf{0.78} & \textbf{0.76} & \textbf{0.77} \\ \hline
\end{tabular}
\caption{Comparison of different models on the plot-to-table conversion task on the PlotQA dataset.}
\label{tab:f1-score}
\end{table}

\noindent \textbf{Ablation Studies:} Comparing rows (a) and (b) with (c) of Table~\ref{tab:plotnet_map}, we observe that (1) adding linking module allows us to handle longer textual elements (\textit{e.g.}, AP for plot-title improved from 0.00\% to 99.44\%), and (2) adding neighbourhood information using AN-ROI leads to a significant improvement in the performance (\textit{e.g.}, mAP@0.9 improved from 83.89\% to 91.77\%).
Rows (d) to (i) suggest that when we use any of the IOU-based loss functions, the mAP@0.9 is lower than what we obtain by using only the smooth $\ell_1$-loss (row (c)). However, among the IOU-based loss functions, FIOU gives the best performance. Further, using our custom loss (last row), we get the best performance with an overall mAP@0.9 of $93.44\%$.

\begin{figure}[t!]
\includegraphics[scale=0.22]{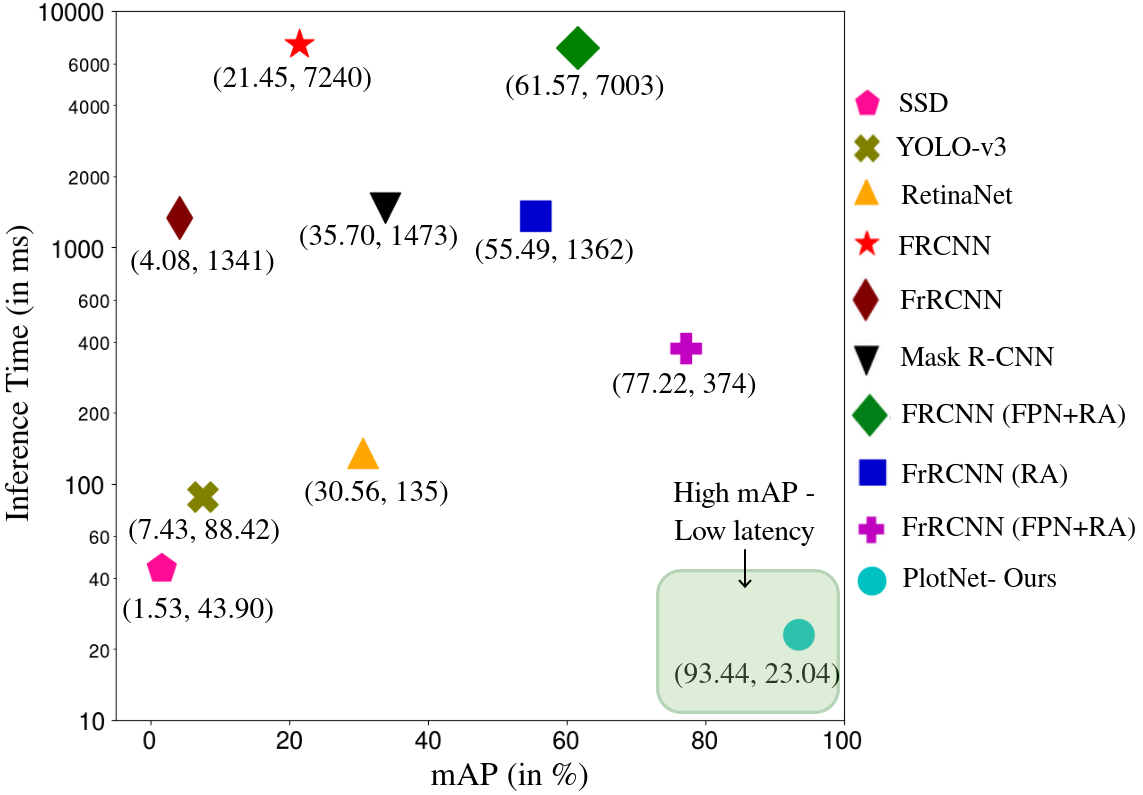}
\caption{mAP (in \%) v/s Inference Time per image (in ms) for different object detection models on PlotQA at an IOU setting of 0.9. (x, y) represents the tuple (mAP, time).
}
\label{fig:latency}
\end{figure}

\noindent \textbf{Comparison to other models:} In Fig.~\ref{fig:latency}, we compare the mAP@0.9 and inference time of different models. We observe that PlotNet lies in the most favorable regime, \textit{i.e.}, high mAP and low latency. In particular, PlotNet has the smallest inference time, beating one-stage detectors. Further, it improves upon its closest competitor (FrRCNN (FPN+RA)) by 16.22 absolute points in mAP. 
We also refer to Fig.~\ref{fig:proposals_loss_comparison_plotnetv7}(b) which shows that PlotNet detects accurate boxes. We note that this example is representative of the overall performance. Amongst individual classes, PlotNet majorly improves the accuracy on plot-titles which have long texts. We attribute this to combining simple region proposals with an explicit linking method. The improved accuracy of PlotNet on small objects like dot-line can be attributed to the additional neighbouring information present in each ROI feature.

\noindent \textbf{Extrinsic Evaluation:} %
Once the objects in a plot are accurately detected they can be used for inferences in a downstream task. For example, the data encoded in the plot can be represented as a structured table and then QA can be performed on this structured table. We follow the same procedure as outlined in \cite{Methani_2020_WACV} to construct a structured table from the objects identified in the plot.
The quality of the generated structured table clearly depends on accurate localisation and classification of objects in the plot. The original PlotQA dataset also provides the gold standard structured tables associated with each plot. We can compute the F1-score by comparing the tables generated after object detection with the ground truth tables. We report these numbers in Table \ref{tab:f1-score}. We observe that PlotNet improves the F1-score by a significant 20 points w.r.t its closest competitor - Faster R-CNN (FPN+RA). This signifies that the improved results with PlotNet enable automated reasoning with plots.

\section{Conclusion}   
\label{sec:conclusion}
Existing object detection networks do not work for scientific plots - they have very low accuracy at the high IOU values required for reasoning over plots. 
Our proposed PlotNet makes a series of contributions across region proposal, model design, feature extraction, and loss function. 
These contributions together give a significant improvement of 16.22 points at mAP@0.9 IOU. 
PlotNet is also much faster with a 16x speedup in comparison with existing networks, including one-stage detectors.
On the extrinsic challenging task of plot-to-table, PlotNet provides an improvement of 20 points in the F1 score.
These significant results enable further exploration of automated reasoning over plots.

\section{Acknowledgement}  
We are extremely thankful to Google for funding this work. Such extensive experimentation would not have been possible without their invaluable support.

{\small
\bibliography{main}
}

\section{Supplementary}
The supplementary material is organised as follows: \\
We first briefly introduce the different approaches for object detection and discuss the key ideas of the existing CNN-based object detection models.
We then explain the motivation behind our proposed CV-based region proposal method and the Focal IOU (FIOU) loss function used in PlotNet (our proposed model).
We also discuss the results and observations of hyper-parameter tuning experiments. Lastly, we explain the extrinsic task of plot-to-table conversion and exemplify the idea with the help of a few examples.

\section{Existing Models for Object Detection}
\label{sec:existing_models}
The goal of object detection is two-fold: (i) identify instances of a predefined class, (ii) identify its location with a bounding box. Over the last few years, the traditional complex ensemble object detection systems have been replaced by convolutional neural networks. There are two different approaches for doing object detection: (i) we can either make a fixed number of predictions on the feature map or the image itself by dividing it into $S \times S$ grid and then predict the class of the object present and the regressed co-ordinates of the bounding box; or (ii) leverage a separate proposal network to find relevant objects and then use a second network to fine-tune the proposals and output a final prediction. Based on the above criteria, we have one-stage and two-stage detectors.

In the following paragraphs, we summarise various CNN-based models along with their key insights.

\noindent\textbf{R-CNN} \cite{DBLP:conf/cvpr/GirshickDDM14} was the first CNN-based object detector. 
It uses selective search (SS) \cite{DBLP:journals/ijcv/UijlingsSGS13} which combines the best of exhaustive search and segmentation to identify about $2,000$ candidate region proposals.
Each region proposal is independently classified based on features extracted from the cropped region of that proposal. \\
\noindent\textbf{Fast R-CNN} \cite{DBLP:conf/iccv/Girshick15} speeds-up R-CNN by computing the features for all proposals jointly from an intermediate layer of a CNN. 
It also introduced ROIPool which warps the variable sized proposals into a fixed-size before classification and regression tasks on fully connected layers. \\
\noindent\textbf{Faster R-CNN} \cite{DBLP:conf/nips/RenHGS15} replaced an external proposal method with a Region Proposal Network (RPN) which learns to predict proposals on pre-defined anchors of different sizes on different parts of the images. \\
\noindent\textbf{Mask R-CNN} \cite{DBLP:conf/iccv/HeGDG17} was proposed for instance segmentation and uses two stage approach for detecting and classifying objects similar to Faster R-CNN. The authors observed that ROIPool leads to harsh quantisations of the proposed regions and hence they introduce \textbf{ROIAlign} which uses bilinear interpolation to calculate the feature map values within a scaled down size of the proposal.\\
\noindent\textbf{Single Shot MultiBox Detector (SSD)} \cite{DBLP:conf/eccv/LiuAESRFB16} is a one-stage detector which substitutes the RPN by multiple object predictions at different pre-identified cells.
These proposals are identified on feature maps of different resolutions to detect objects of different scales. 
This speeds up detection but at the cost of accuracy, relative to Faster R-CNN.\\
\noindent\textbf{RetinaNet} \cite{DBLP:conf/iccv/LinGGHD17} uses a fixed anchor based proposal generation technique on each layer of a Feature Pyramid Network (\textbf{FPN}).
FPN is a feature extractor designed for multi-scale object detection with both accuracy and speed considerations. 
It combines low-resolution, semantically strong features with high-resolution, semantically weak features via a top-down pathway and lateral connections. 
This generates a huge number ($\sim200$K) of proposals leading to class imbalance challenge which is addressed with a custom loss function called focal loss.\\
\noindent\textbf{YOLO-v3} \cite{DBLP:journals/corr/abs-1804-02767} uses bounding box priors with varying aspect ratios identified by K-means clustering on all bounding boxes in the training dataset. 
It addresses the challenge of class imbalance by first predicting the probability whether the object is present and then predicting the object's class conditional probability.

\noindent In summary, one-stage detectors trade accuracy for real-time processing speed whereas two-stage detectors have higher accuracy as the proposals undergo a two-stage filtering and regression, first through RPN and then through classification and regression heads. 
However, these models are compute intensive as they use ResNet-50 as their feature extractor.

\section{PlotNet: CV-based Region Proposal}
\begin{figure*}[t]
\centering
\includegraphics[scale=0.57]{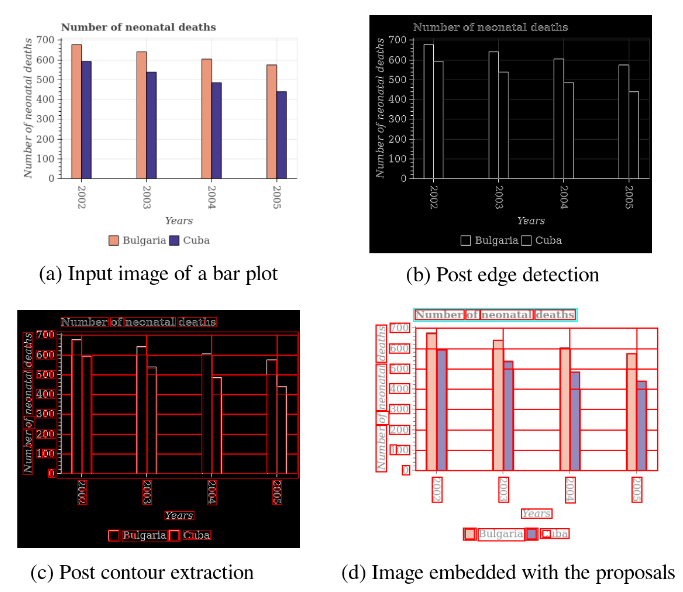}
\caption{Various steps involved in our CV-based region proposal generation method. Proposals generated by our CV method is shown in red. Note that the longer textual elements such as y-axis label, plot-title (ground-truth box shown in cyan) gets split across multiple proposals which can be merged by the linking module of PlotNet.} 
\label{fig:proposals}
\end{figure*}

In this section, we elaborate the motivation behind our CV-based region proposal method. 
On an average, there are $\sim30$ objects present in a scientific plot and using any anchor-based region proposal techniques which proposes $\sim100K$ boxes, is clearly an overkill.
In the plot shown in Fig.~\ref{fig:proposals}a, we can see that in the upper strip of the image, to capture the plot-title, we need a long rectangular box. Therefore, it is unnecessary to try and fit anchors of all possible sizes at every feature point of this image.
Given the simplicity of the objects found in scientific plots, we leverage the traditional CV-based methods to generate an initial set of proposals. To this end, we tried different approaches:
i) Color-based segmentation method: It draws a bounding box for the objects having the same color. This approach fails to work when the neighbouring objects have colors of a similar shade.
ii) Connected components method: It draws a bounding box for objects that share a common edge. This approach fails to work in the case of adjacent bars and results in a single merged proposal. For \textit{e.g.}, in Fig.~\ref{fig:proposals}a, a traditional connected component algorithm will merge the bars denoting Bulgaria and Cuba into a single proposal, resulting in a loss of information.
Therefore, we use a sequence of edge-detection algorithms to generate our final proposals.

We first use a Laplacian edge detector on the image of the plot, which uses Laplacian kernels to approximate a second derivative measurement on the image. This helps us to detect the edges in the image by highlighting regions showing a rapid change in intensity.
\begin{table*}[ht!]
\centering
\begin{tabular}{@{}clcccccccccccc@{}}
\toprule
 &  & \multicolumn{10}{c}{\textbf{IOU @0.9}} & \multicolumn{1}{l}{\textbf{@0.75}} & \multicolumn{1}{l}{\textbf{@0.5}} \\ \midrule
\textbf{S.L} & \multicolumn{1}{c}{\textbf{\begin{tabular}[c]{@{}c@{}}Feature\\ Extractor\end{tabular}}} & \textbf{bar} & \textbf{\begin{tabular}[c]{@{}c@{}}dot-\\ line\end{tabular}} & \textbf{\begin{tabular}[c]{@{}c@{}}legend\\ label\end{tabular}} & \textbf{\begin{tabular}[c]{@{}c@{}}legend\\ preview\end{tabular}} & \textbf{\begin{tabular}[c]{@{}c@{}}plot\\ title\end{tabular}} & \textbf{\begin{tabular}[c]{@{}c@{}}x-axis\\ label\end{tabular}} & \textbf{\begin{tabular}[c]{@{}c@{}}x-axis\\ ticks\end{tabular}} & \textbf{\begin{tabular}[c]{@{}c@{}}y-axis\\ label\end{tabular}} & \textbf{\begin{tabular}[c]{@{}c@{}}y-axis\\ ticks\end{tabular}} & \textbf{mAP} & \textbf{mAP} & \textbf{mAP} \\ \cmidrule(l){1-14} 
(a) & R-10 & 89.46 & 37.63 & 93.64 & 78.12 & 95.22 & 94.06 & 88.10 & 66.95 & 83.86 & 80.78 & 96.74 & 97.57 \\
(b) & R-22 & 91.37 & 24.08 & 97.03 & 81.01 & 98.79 & 90.47 & 81.99 & 51.79 & 47.92 & 73.83 & 97.01 & 98.08 \\
(c) & R-50 & 87.64 & 15.72 & 74.57 & 41.87 & 98.92 & 81.60 & 54.21 & 43.35 & 35.67 & 59.28 & 93.91 & 97.67 \\
(d) & R-22 FPN & 91.02 & 31.69 & 97.08 & 81.57 & 99.36 & 96.06 & 85.33 & 82.00 & 90.95 & \textbf{83.89} & \textbf{97.21} & \textbf{98.11} \\
(e) & R-50 FPN & 90.77 & 5.12 & 95.58 & 80.72 & 99.16 & 94.79 & 76.83 & 65.56 & 58.17 & 74.08 & 94.09 & 97.67 \\ \bottomrule
\end{tabular}
\caption{Hyperparameter tuning for feature extractor backbone of PlotNet: Comparison of different variants of PlotNet on the PlotQA dataset by varying the number of layers in the ResNet(R)-50 architecture with mAP scores (in \%) at IOUs of 0.9, 0.75, and 0.5.}
\label{tab:feature_extractor_tuning}
\end{table*}
\begin{table*}[ht!]
\centering
\begin{tabular}{@{}clcccccccccccc@{}}
\toprule
 &  & \multicolumn{10}{c}{\textbf{IOU @0.9}} & \multicolumn{1}{l}{\textbf{@0.75}} & \multicolumn{1}{l}{\textbf{@0.5}} \\ \midrule
\textbf{S.L} & \multicolumn{1}{c}{$\gamma$} & \textbf{bar} & \textbf{\begin{tabular}[c]{@{}c@{}}dot-\\ line\end{tabular}} & \textbf{\begin{tabular}[c]{@{}c@{}}legend\\ label\end{tabular}} & \textbf{\begin{tabular}[c]{@{}c@{}}legend\\ preview\end{tabular}} & \textbf{\begin{tabular}[c]{@{}c@{}}plot\\ title\end{tabular}} & \textbf{\begin{tabular}[c]{@{}c@{}}x-axis\\ label\end{tabular}} & \textbf{\begin{tabular}[c]{@{}c@{}}x-axis\\ ticks\end{tabular}} & \textbf{\begin{tabular}[c]{@{}c@{}}y-axis\\ label\end{tabular}} & \textbf{\begin{tabular}[c]{@{}c@{}}y-axis\\ ticks\end{tabular}} & \textbf{mAP} & \textbf{mAP} & \textbf{mAP} \\ \cmidrule(l){1-14} 
(a) & 0.85 & 92.07 & 56.89 & 92.50 & 93.94 & 99.55 & 73.97 & 88.80 & 83.80 & 58.36 & 82.80 & 97.30 & 98.28 \\
(b) & 1.00 & 92.03 & 60.37 & 95.23 & 95.11 & 98.85 & 97.02 & 88.88 & 96.50 & 81.45 & 89.49 & 97.27 & 98.13 \\
(c) & 1.25 & 92.14 & 55.59 & 93.69 & 95.57 & 99.43 & 90.24 & 84.96 & 91.33 & 58.70 & 84.63 & 96.50 & 98.22 \\
(d) & 1.50 & 92.06 & 51.74 & 94.84 & 94.98 & 94.84 & 93.31 & 89.53 & 93.79 & 55.31 & 84.49 & 95.94 & 97.83 \\
(e) & 1.75 & 91.99 & 56.72 & 95.70 & 95.67 & 99.49 & 96.43 & 92.12 & 95.34 & 87.63 & 90.14 & 96.98 & 98.22 \\
(f) & 2.00 & 91.88 & 61.44 & 96.44 & 95.58 & 99.52 & 97.19 & 90.64 & 97.55 & 87.66 & \textbf{90.88} & \textbf{97.36} & \textbf{98.31} \\
(g) & 3.00 & 92.27 & 65.13 & 93.99 & 95.28 & 99.26 & 91.31 & 89.00 & 95.39 & 65.82 & 87.55 & 97.33 & 98.27 \\
(h) & 4.00 & 92.50 & 59.76 & 97.03 & 86.39 & 99.31 & 96.74 & 89.70 & 96.94 & 92.89 & 90.14 & 95.94 & 97.07 \\ \bottomrule
\end{tabular}
\caption{Hyperparameter tuning for $\gamma$ in FIOU Loss: Comparison of PlotNet (proposed model) on the PlotQA dataset with the FIOU loss (proposed loss function) for different values of $\gamma$ with mAP scores (in \%) at IOUs of 0.9, 0.75, and 0.5.}
\label{tab:hyperparam_tuning_gamma}
\end{table*}
Once the edges are identified, we extract contours by joining all the consecutive points along the boundary of the edges, based on uniform colour and intensity.
Lastly, we convert these contours into a bounding box by finding the minimal area up-right bounding rectangle for each of the identified contours. These boxes serve as regions of interest (ROI) which are passed as input to the network.
Our proposed method is $\sim150$ times faster than traditional selective search simply because applying Laplacian kernels requires very few arithmetic operations. The output of each stage is shown in Fig.~\ref{fig:proposals}.
It can be observed that our region proposal method has a high recall as it detects all the objects present in the image.

\begin{figure}[ht!]
\centering
\includegraphics[scale=0.32]{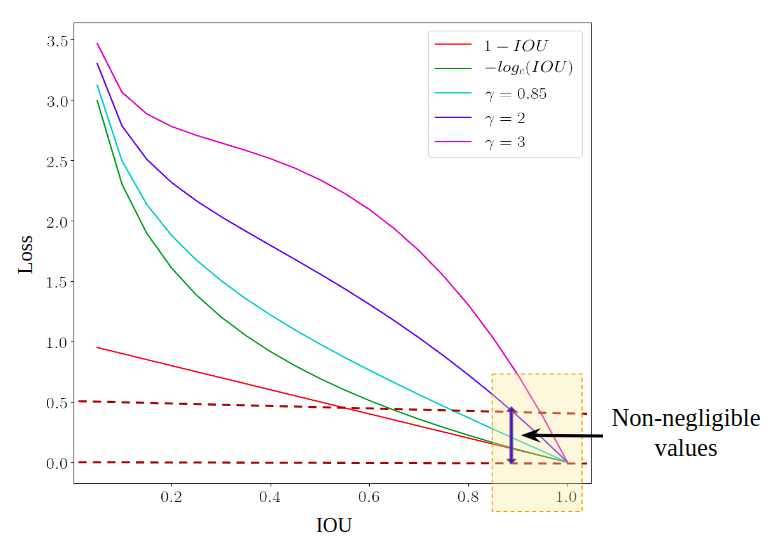}
\caption{Comparison of different loss functions at varying IOUs. Note that FIOU with varying values of $\gamma$ gives non-negligible values at high IOUs.} 
\label{fig:loss_comparison}
\end{figure}

\section{PlotNet: Focal IOU (FIOU) loss}
In this section, we compare our proposed Focal IOU loss with other IOU-based loss functions. As shown in Fig.~\ref{fig:loss_comparison}, existing IOU-based loss functions have negligible values if the IOU overlap between the predicted box and the ground-truth box is reasonably
high (say, $> 0.8$). As discussed in the main paper, while this is acceptable for objects in natural images, it is not suitable for objects in scientific plots that require stricter localization.
To this end, we propose a custom loss function that gives non-negligible values at high IOUs. 
The idea is similar to Focal Loss \cite{DBLP:conf/iccv/LinGGHD17} that scales the contribution of each sample to the loss based on the classification error, \textit{i.e.}, higher the classification error, lesser is the contribution of that sample to the loss function. Focal IOU (FIOU) loss function follows a similar principle. In particular, it increases the weight of samples which are already well localized so that their contribution to the loss is non-negligible.
FIOU thus focuses on higher IOU values by dynamically scaling the $-\log(IOU)$ which pushes the network towards learning very tight bounding boxes.

\section{Hyper-parameter Tuning Experiments}

\begin{figure*}[ht!]
\centering
\begin{subfigure}{.45\textwidth}
\centering
\includegraphics[scale=0.35]{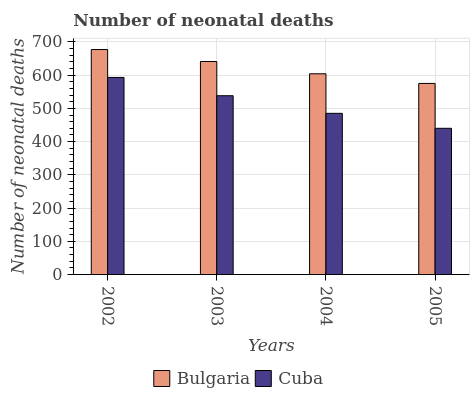}
\caption{Input bar plot image}
\label{fig:input_image}
\end{subfigure}
\hspace{0.5cm}
\begin{subfigure}{.45\textwidth}
\centering
\includegraphics[scale=0.30]{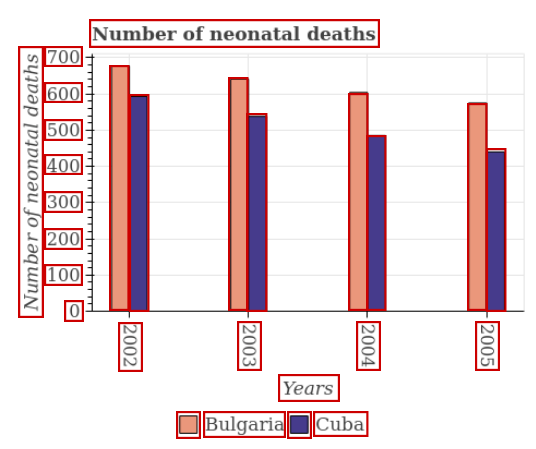}
\caption{Predicted bounding boxes}
\label{fig:predicted}
\end{subfigure}
\hspace{0.5cm}
\begin{subfigure}{.45\textwidth}
\centering
\includegraphics[scale=0.45]{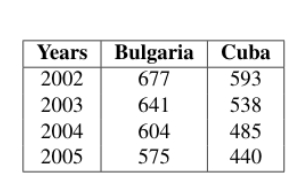}
\caption{Ground-truth table}
\label{fig:gt_table}
\end{subfigure}
\hspace{0.5cm}
\begin{subfigure}{.45\textwidth}
\centering
\includegraphics[scale=0.45]{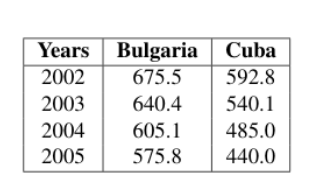}
\caption{Generated table}
\label{fig:generated_table}
\end{subfigure}
\caption{Sample table generation using PlotNet's prediction on an image of a bar plot.}
\label{fig:plotnet_prediction_1}
\end{figure*}

\begin{figure*}[ht!]
\centering
\begin{subfigure}{.45\textwidth}
\centering
\includegraphics[scale=0.40]{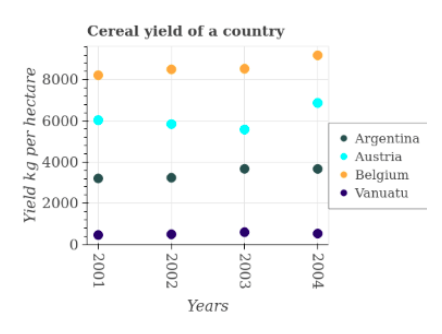}
\caption{Input dot-line plot image}
\label{fig:input_image_2}
\end{subfigure}
\hspace{0.5cm}
\begin{subfigure}{.45\textwidth}
\centering
\includegraphics[scale=0.40]{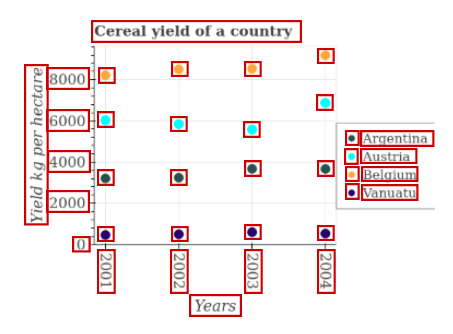}
\caption{Predicted bounding boxes}
\label{fig:predicted_2}
\end{subfigure}
\hspace{0.5cm}
\begin{subfigure}{.45\textwidth}
\centering
\includegraphics[scale=0.40]{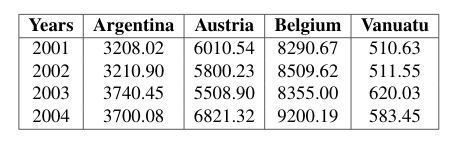}
\caption{Ground-truth table}
\label{fig:gt_table_2}
\end{subfigure}
\hspace{0.5cm}
\begin{subfigure}{.45\textwidth}
\centering
\includegraphics[scale=0.40]{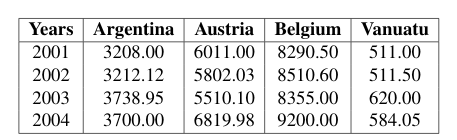}
\caption{Generated table}
\label{fig:generated_table_2}
\end{subfigure}
\caption{Sample table generation from PlotNet's prediction on an image of a dot-line plot.}
\label{fig:plotnet_prediction_2}
\end{figure*}

\noindent \textbf{Feature Extractor: }
We compare the performance of PlotNet by varying the number of layers in the ResNet-$50$ feature extractor backbone. Note that for the following experiment, we have chosen to omit the AN-ROI layer from the architecture of PlotNet.

From Table~\ref{tab:feature_extractor_tuning}, we can observe that by using R-$22$ with FPN, \textit{i.e.},  ResNet-22 with FPN as the feature extractor gives the highest mAP throughout at $0.9$, $0.75$ and $0.5$ IOUs (row (d)). Hence we choose the same in all of our further experiments. We observe that even a very shallow backbone network of $10$ layers (R-$10$) achieves the second-best performance (row (a)) which supports the fact that bulky networks are not necessary for detecting simple objects like the ones present in scientific plots. The performance degrades (especially for small objects like dot-line and legend-preview) when we use too deep bulky networks such as R-$50$ with or without FPN (rows (c) and (e)) which suggests that features for tiny objects are lost. Hence, they are either misclassified or incorrectly localized.

\noindent \textbf{Tuning $\gamma$ in FIOU loss: } We compare the performance of PlotNet with the IOU-based custom loss function for different values of the scaling factor $\gamma$.

From Table~\ref{tab:hyperparam_tuning_gamma}, we see that \textbf{$\gamma=2.00$} gives the highest mAP throughout at $0.9$, $0.75$ and $0.5$ IOUs (row (f)), and hence we choose the same in all of our further experiments. We also observe that as $\gamma$ increases from $0.85$ to $4.00$, AP for the textual elements such as legend-label, x, y ticks and labels, increases, and then drops while achieving its peak value at $\gamma=2.00$.

\section{Plot-to-Table Conversion}
As discussed in the main paper, detecting the objects in a scientific plot can be used to extract the underlying data that the plot visually represents. To this extent, we take a step forward and describe the process of table generation from plots using the example given in Fig.\ref{fig:plotnet_prediction_1}. We have used the rules introduced in PlotQA \cite{Methani_2020_WACV} for the same.
Specifically, referring to the detected elements in the Fig.~\ref{fig:predicted} and the generated table shown in Fig.~\ref{fig:generated_table}, we see that tick-labels (2002, 2003, 2004, and 2005) on the $x$-axis correspond to the rows of the table whereas the different labels (Cuba and Bulgaria) listed in the legend correspond to the columns of the table. The $i$, $j$-th cell of the table denotes the value corresponding to the $i$-th x-tick and the $j$-th legend-label. The values of all the textual elements such as plot-title, legend-labels, and the $x,y$-tick labels can be obtained from a pre-trained Optical Character Recognition (OCR) engine\footnote{https://github.com/tesseract-ocr/tesseract}. Keeping in mind the spatial structure of scientific plots, the mapping of legend-label to legend-preview, $x,y$ ticks to corresponding $x,y$ labels and bounding boxes of bars to their corresponding $x$-ticks and legend-labels, is carried out. The height of each bar is extracted by using the top-left and bottom-right bounding box coordinates and the immediate $y$-tick label above or below that height. Finally, the value of each bar is interpolated based on the previously extracted bounding ticks. 

We can follow the above method to extract the underlying data from a dot-line plot given in Fig.\ref{fig:input_image_2} into a similar looking table, as shown in Fig.\ref{fig:generated_table_2}.

On comparing the ground-truth tables shown in Fig.\ref{fig:gt_table} and \ref{fig:gt_table_2} with the corresponding generated tables shown in Fig.\ref{fig:generated_table} and \ref{fig:generated_table_2}, we can see both the generated tables are very close to their respective ground-truth tables \textit{i.e.}, almost all the numeric values in the generated tables lie within $2\%$ of the corresponding values in the ground-truth table. This suggests that PlotNet can detect all the objects that are present in scientific plots with very high localization accuracy.

\end{document}